\documentclass[letterpaper, 10 pt, conference]{packages/ieeeconf}
\IEEEoverridecommandlockouts

\usepackage{packages/algorithm}
\usepackage{packages/algorithmic}

\usepackage[compress]{cite}
\makeatletter
\let\NAT@parse\undefined
\makeatother
\usepackage[colorlinks=true]{hyperref}
\usepackage[table]{xcolor} % Boxed equations http://tex.stackexchange.com/a/14298
\usepackage[section]{placeins} %prevent figure foating 
\usepackage{graphics} % for pdf, bitmapped graphics files
\usepackage{graphicx}
\usepackage{cuted}
\usepackage{amsmath} % assumes amsmath package installed
\usepackage{mathtools} % left align
\usepackage{todonotes}
\usepackage{amssymb}  % assumes amsmath package installed
\usepackage{makeidx} % allows index generation
\usepackage{graphicx} % standard LaTeX graphics tool
\graphicspath{{figures/}}
\usepackage{multicol} % used for the two-column index
\usepackage[bottom]{footmisc} % places footnotes at page bottom
\usepackage[utf8]{inputenc} % allows for new lines
\usepackage{multirow}
\usepackage{capt-of}
\usepackage{booktabs} % for \toprule tables
\usepackage{adjustbox}
\usepackage{caption}
\usepackage{makecell} % https://tex.stackexchange.com/a/202893
\usepackage{subcaption} % side by side figures
\usepackage{tikz}
\usepackage{booktabs} % add this to your preamble
\usetikzlibrary{positioning}
\usetikzlibrary{shapes,snakes}
\usepackage{bm} % allows use of \bm command in math mode
\usepackage{float} % allow using of [H] tag on "figure"s
\usepackage{color} % color text package
\usepackage{empheq} % Boxed equations http://tex.stackexchange.com/a/14298
\usepackage{threeparttable} % Table bottom note!
\usepackage{xcolor}
\usepackage{xurl} % Allow url to break line more freely.
\usepackage{tcolorbox}
\usepackage{pifont}   % for \ding{51} and \ding{55}
\usepackage{float}
\usepackage{todonotes}

\DeclareMathSymbol{\shortminus}{\mathbin}{AMSa}{"39}
\usepackage{paralist}

\usepackage{mathtools}
\DeclarePairedDelimiterX{\norm}[1]{\lVert}{\rVert}{#1}

\allowdisplaybreaks
\renewcommand{\baselinestretch}{0.97}
\begin{document}

\title{\LARGE \bf
% Monocular Visual-Inertial Odometry with Metric Dense Depth Estimation
VIMD: Monocular Visual-Inertial Motion and Depth Estimation
}

\author{Saimouli Katragadda and Guoquan Huang
\thanks{This work was partially supported by 
the University of Delaware (UD) College of Engineering, 
and Google ARCore.
}
\thanks{$^{1}$The authors are with the Robot Perception and Navigation Group (RPNG), University of Delaware, Newark, DE 19716.
Email: {\tt\small \{saimouli, ghuang\}@udel.edu}.}
}

% $^{2}$ Google AR Core Email: \{chaoguo, mingyangli\}@google.com}

% Create ieeeconf title page
\maketitle
% \vspace*{-29\baselineskip}

% % --- Full-width teaser banner (no float) ---
% \begingroup
%   %\setlength{\stripsep}{0pt plus 1pt minus 1pt}% cuted strip padding
%   %\setlength{\abovecaptionskip}{2pt}
%   %\setlength{\belowcaptionskip}{0pt}
%   \begin{strip}
%     \centering
%     \includegraphics[width=\textwidth]{figures/teaser_depth.png}% or teaser_depth.png
%     \captionof{figure}{Outdoor and indoor dense depth metric mapping}
%     \label{fig:teaser}
%   \end{strip}
% \endgroup

% \thispagestyle{empty}
% \pagestyle{empty}
% \setcounter{page}{1}
% \pagestyle{empty}

% Main sections

\begin{abstract}
Accurate and efficient dense metric depth estimation is crucial for 3D visual perception in robotics and XR.
In this paper, we develop a monocular visual-inertial motion and depth (VIMD) learning framework to estimate dense metric depth by leveraging accurate and efficient MSCKF-based monocular visual-inertial 3D motion tracking. 
At the core the proposed VIMD is to exploit  multi-view information to iteratively refine per-pixel scale, 
instead of globally fitting an invariant affine model as in the prior work.
The VIMD framework is highly modular, making it compatible with a variety of existing depth estimation backbones.
We conduct extensive evaluations on the TartanAir and VOID datasets and demonstrate its zero-shot generalization capabilities on the AR Table dataset. 
Our results show that VIMD achieves exceptional accuracy and robustness, even with extremely sparse points—as few as 10-20 metric depth points per image. 
This makes the proposed VIMD a practical solution for deployment in resource-constrained settings, 
while its robust performance and strong generalization capabilities offer significant potential across a wide range of scenarios.

\end{abstract}

%=====================================================
%=====================================================
%=====================================================
\section{Introduction}

For applications like robotics and extended reality (XR), accurate and efficient metric dense depth estimation is critical for 3D visual perception, which is essential for tasks such as obstacle avoidance and motion planning. 
Monocular methods, which estimate depth from a single RGB image, are particularly appealing because they use a compact, inexpensive, and common camera. However, purely monocular vision suffers from an inherent scale ambiguity: it can capture the relative shape of a scene but cannot determine the true distances to objects.
Knowing exactly how far an obstacle is from the camera can be the difference between safe passage and collision. 
Integrating inertial data can (partially) resolve this ambiguity.
As most mobile devices and robots are already equipped with an IMU,
visual–inertial odometry (VIO) or SLAM prevails but typically yields only sparse metric depth from a small set of tracked landmarks~\cite{Huang2019ICRA},
which alone cannot provide the dense metric depth map required in safety-critical operations.

\begin{figure}
    \centering
   % \vspace{-2mm} 
    \includegraphics[width=1.0\linewidth, trim={1mm 1mm 1mm 2mm}, clip]{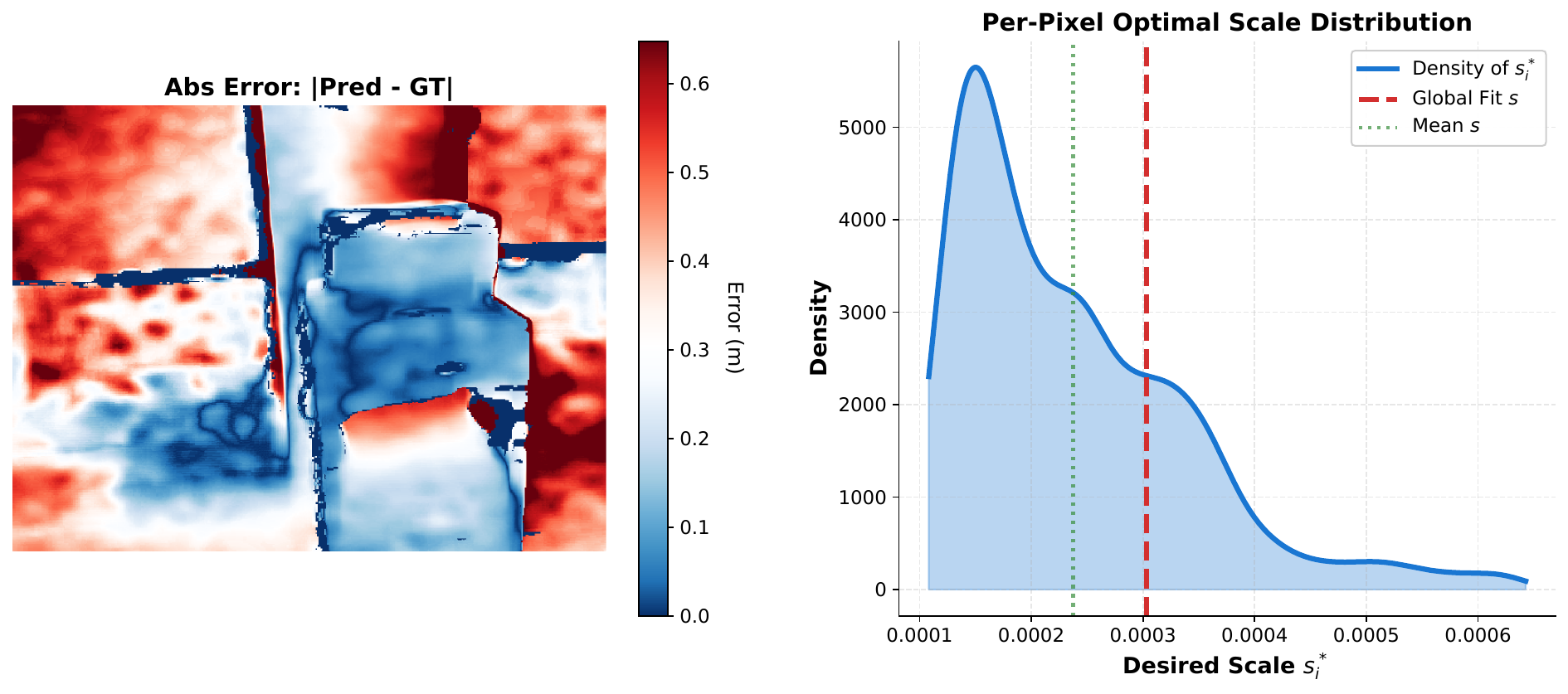}
    %\vspace{-2mm}
    \caption{
    A motivation result: A single global affine model fails for metric depth recovery from monocular predictions. Left: depth error after per-frame global scale alignment. Right: per-pixel ideal scale from ground truth, revealing strong spatial variation.
   % A single global affine model is insufficient for metric depth recovery from monocular predictions. 
   % (Left: error map of depth aligned using per-frame global scale fitting; 
   % Right: per-pixel ideal scale computed from ground-truth depth and the globally aligned prediction for the same frame).
   %  The strong spatial variation in scale indicates that monocular depth predictions are not globally affine with respect to metric depth.
    }
    \label{fig:intro_motiv}
    %\vspace{-2mm}
\end{figure}

Although recent advances in monocular depth prediction achieve high-quality relative depth estimation \cite{Ranftl2021_ICCV, Birkl2023_arxiv}, these methods still lack the ability to produce depth estimates with absolute metric scale.
A common approach assumes an invariant affine relationship and aligns the predicted depth to sparse visual–inertial odometry (VIO) depth measurements via per-frame least-squares fitting of a global scale and offset (e.g., see \cite{Merrill2023RSS, Merrill2024_IJRR}). While simple and computationally efficient, this approach implicitly assumes that a single global affine transformation is sufficient to correct the predicted depth across the entire image.
% However, as shown in Fig. \ref{fig:intro_motiv}, the fitted scale and offset vary significantly across frames—particularly the offset, which exhibits much larger fluctuations than the scale. 
% This temporal instability indicates that the predicted relative depths are not truly affine with respect to the metric depths; that is, they cannot be accurately modeled by a single global scale and offset per frame without unmodeled errors. 

As shown in Fig.~\ref{fig:intro_motiv}, globally aligned depth (left) exhibits noticeable errors that increase as fewer sparse depth points are available (see Tab.~\ref{tab:combined_metrics_selected_reduce}). To further analyze this, we compute the ideal scale at each pixel using ground-truth depth and the globally aligned depth. The resulting scale map (right) shows strong spatial variation, indicating that predicted and metric depth are not globally affine. As a result, enforcing a single global scale (or scale and offset) introduces systematic, spatially varying errors that per-frame fitting cannot correct.

Beyond the unreliability of per-frame global fitting, the optimization becomes poorly conditioned when sparse depth points are low. This leads to increased sensitivity of the fitted parameters and potential error amplification, particularly in low-density regions. Furthermore, it has been empirically observed that jointly predicting an offset along with scale can degrade depth accuracy~\cite{Wofk2023_ICRA}.

% From  Fig. \ref{fig:intro_motiv}, it is clear that the offset exhibits much higher variance across frames than the scale, implying that offsets may also vary more significantly across pixels within a single frame (due to factors like viewpoint changes, occlusions, or non-uniform scene structure), thus making them harder to learn reliably at the per-pixel level compared to the more stable scale.
%

% Motivated by these observations, we propose 
% to exploit multi-view constraints and learn to iteratively optimize a per-pixel scale, moving beyond global affine parameter fitting. Specifically, we introduce a monocular Visual–Inertial Motion and Depth (VIMD) framework that leverages an accurate and efficient MSCKF-based visual–inertial motion estimation module to establish multi-view geometric constraints, and then refines dense depth predictions along with their uncertainty.
% As a result, the proposed VIMD improves accuracy and robustness in low-density depth regions (see Tab.~\ref{tab:combined_metrics_selected}). Fig.~\ref{fig:teaser} illustrates representative results of VIMD in both outdoor and indoor scenes, demonstrating the ability to produce high-quality dense metric depth maps.

\begin{figure*}
    \centering
    \includegraphics[width=0.95\linewidth]{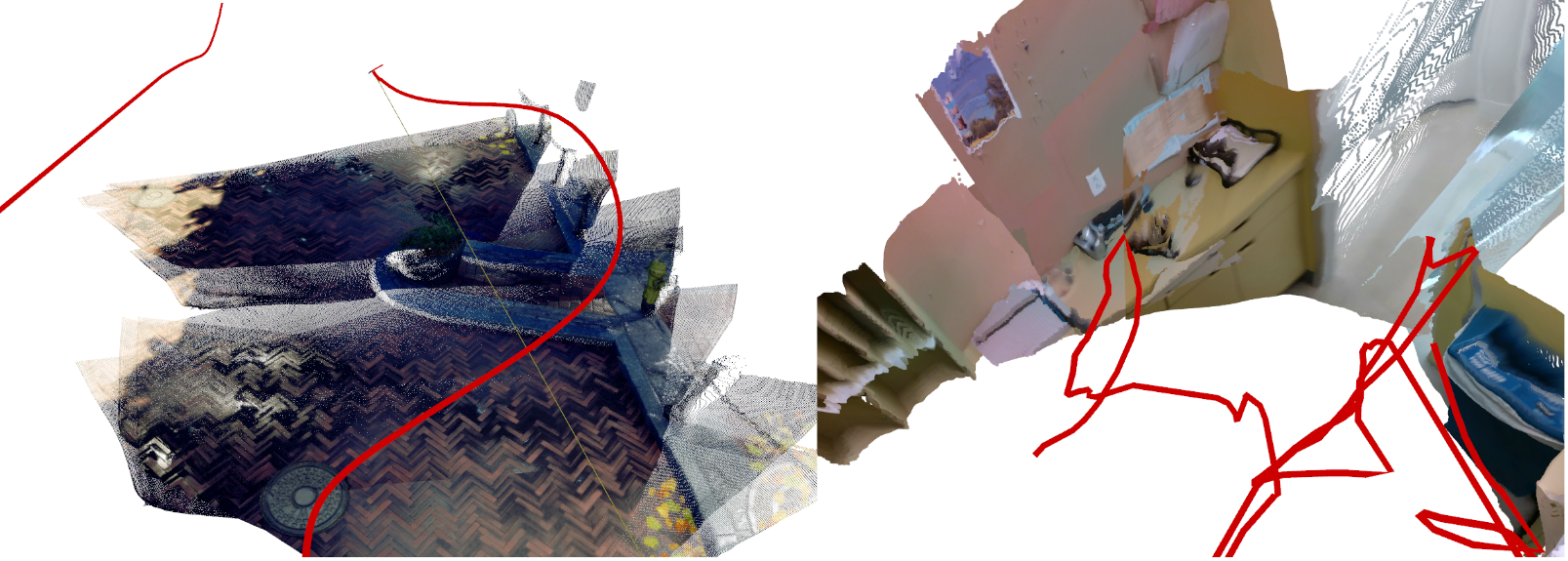}
    \caption{Illustration of the proposed VIMD learning high-quality dense metric depths in both outdoor and indoor scenes.}
    \label{fig:teaser}
\end{figure*}

Motivated by these observations, we investigate how to better leverage sparse VIO depth measurements and camera motion to obtain efficient, robust, and accurate metric depth for 3D spatial perception. 
Our contributions are:
\begin{itemize}
    \item We propose VIMD, a modular monocular visual--inertial framework that leverages multi-view geometric constraints from an efficient MSCKF-based visual--inertial motion estimation to learn to iteratively refine per-pixel scale for dense metric depth estimation, moving beyond per-frame global affine alignment.

\item We develop a lightweight GRU-based iterative depth refinement model, in which 
% the initial metric depth is estimated using the global alignment depth
% which is the metric-aligned depth from the monocular depth estimator, fitted with a global scale and offset via least squares using the sparse depth.
reference frames are warped to the target frame using the predicted depth and VIO poses, and the scale is iteratively refined
using a ConvGRU to predict the final depth and uncertainty. 

    \item We demonstrate that VIMD improves depth accuracy and robustness in low-density depth regions (see Tab.~\ref{tab:combined_metrics_selected_reduce}) and produces high-quality dense metric depth maps in both indoor and outdoor scenes (see Fig.~\ref{fig:teaser}). 
    Extensive experiments further show strong zero-shot generalization and real-time performance.
\end{itemize}

The rest of the paper is structured as follows:
After reviewing the literature in the next section,
we present in detail the proposed metric dense depth learning only from a monocular camera and IMU in Section~\ref{sec:model}.
The extensive experimental results are presented in Section~\ref{sec:exp}.
We conclude the paper in Section~\ref{sec:concl} along with future research directions.

\section{Related Work}

\subsection{Visual–Inertial SLAM and Dense Mapping}

While visual-inertial (VI)-SLAM has matured around tightly coupled odometry and optimization backends, most approaches focus on sparse mapping. 
Kimera~\cite{Rosinol2020_Kimera} augments  state estimation with meshing, building per-keyframe meshes via 2D Delaunay triangulation of sparse tracks and fusing them into a global TSDF representation. 
Similarly, dense aerial mapping in \cite{Yang2017_ICRA} feeds metric VIO poses into a motion-stereo depth module, implicitly transferring the IMU scale into dense maps. 
Recently, DiT-SLAM~\cite{Zhao2022_MDPI} integrates visual, inertial, and dense depth constraints in a tightly coupled graph for dense reconstructions.
Despite these progresses, dense mapping in SLAM pipelines often show limited generalization.

\subsection{Monocular Scaled Depth Learning}

Single-image depth networks such as DPT/MiDaS trained on large-scale, heterogeneous datasets demonstrate strong cross-dataset generalization, but their predictions remain limited to relative (up-to-scale) depth~\cite{Ranftl2021_ICCV, Birkl2023_arxiv, Bhat2023_arxiv}. 
Self-supervised monocular methods such as Monodepth2 further improve scalability by removing the need for ground-truth labels, yet they still inherit the same scale ambiguity \cite{Godard2019_ICCV_monodepth2}. 
To mitigate this limitation, efforts have been taken to incorporate additional cues such as inertial or pose information. 
For instance, in \cite{Fei2019_RAL} gravity estimated from VIO is used to regularize surface orientation and encourage geometrically consistent predictions. 
Recently, \cite{Merrill2023_RSS,Merrill2024_IJRR} propose the use of compact scale and offset parameters to represent depth features, reducing the state size of the estimator while improving inertial initialization and overall efficiency.

\subsection{Latent-Code VI-based Dense Depth Estimation}

Recent approaches have also integrated dense depth estimation directly into VI estimation pipelines through compact latent codes. For example, CodeVIO~\cite{Zuo2021_ICRA} introduced a CVAE-based depth decoder whose low-dimensional code is jointly optimized with the VIO state, enabling dense depth with efficient Jacobians and FEJ updates. 
AB-VINS~\cite{Merrill2024_arxiv} pushed this concept further by replacing explicit 3D landmarks with a compressed ``AB feature'' consisting of a global scale $a$ and bias $b$, and correction terms, which reduces computational cost while producing dense depth. Similarly, DiT-SLAM~\cite{Zhao2022_MDPI} uses low-dimensional depth codes coupled with visual–inertial constraints for dense mapping. 
While effective, these latent-code approaches either constrain depth to low-rank parameterizations or require online optimization of codes, limiting flexibility and generalizability.

\subsection{From Sparse to Dense Depth Completion}

Depth completion leverages sparse VIO depth as anchors for dense prediction. 
VOICED~\cite{Wong2020_RAL} constructs a piecewise-planar scaffold from sparse VIO points and infers dense depth via unsupervised photometric and geometric consistency losses. 
VI-Depth \cite{Wofk2023_ICRA} starts from monocular relative depth, aligns it with sparse VIO depths via a global scale and shift, and then applies a per-pixel refinement network to correct local distortions. 
This yields metrically accurate dense depth and demonstrates strong cross-dataset generalization, but relies heavily on accurate alignment and struggles with sparse or noisy anchors.

% \subsection{Positioning of our approach}
% Our work specifically targets metrically accurate and robust dense depth in monocular visual–inertial settings. Unlike latent-code methods such as CodeVIO, DiT-SLAM, and AB-VINS, we do not optimize a compact depth code inside the estimator, avoiding the need for online code optimization and restrictive parameterizations. Instead, we recover per-pixel metric depth by formulating sparse-to-dense alignment as a learning-based optimization problem guided by multi-view geometric constraints from VIO poses. Compared to VI-Depth and VOICED, our framework fully exploits temporal consistency and multi-view geometry, improving robustness in low-density regions and yielding more reliable depth alongside uncertainty estimates.

\section{Learning Metric Dense Depth}
\label{sec:model}

\begin{figure*}[t]
  \centering
  \begin{subfigure}[t]{0.95\textwidth}
    \centering
    \includegraphics[width=\linewidth]{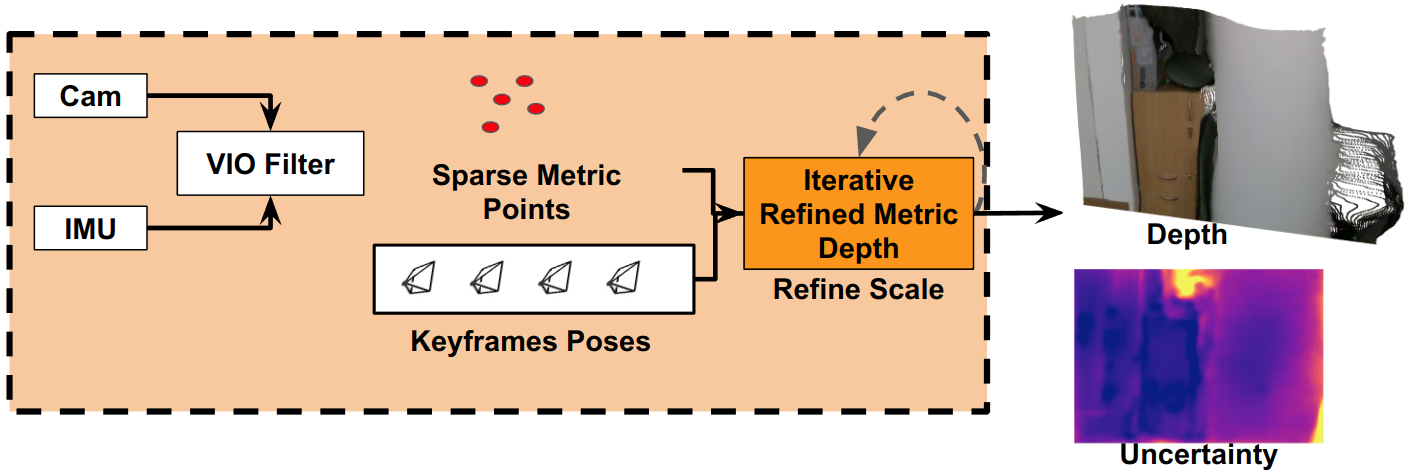}
    \caption{Monocular Visual-Inertial Motion and Metric Depth Estimation Pipeline.}
    \label{fig:pipeline_1}
  \end{subfigure}
  \begin{subfigure}[t]{0.95\textwidth}
    \centering
    \includegraphics[width=\linewidth]{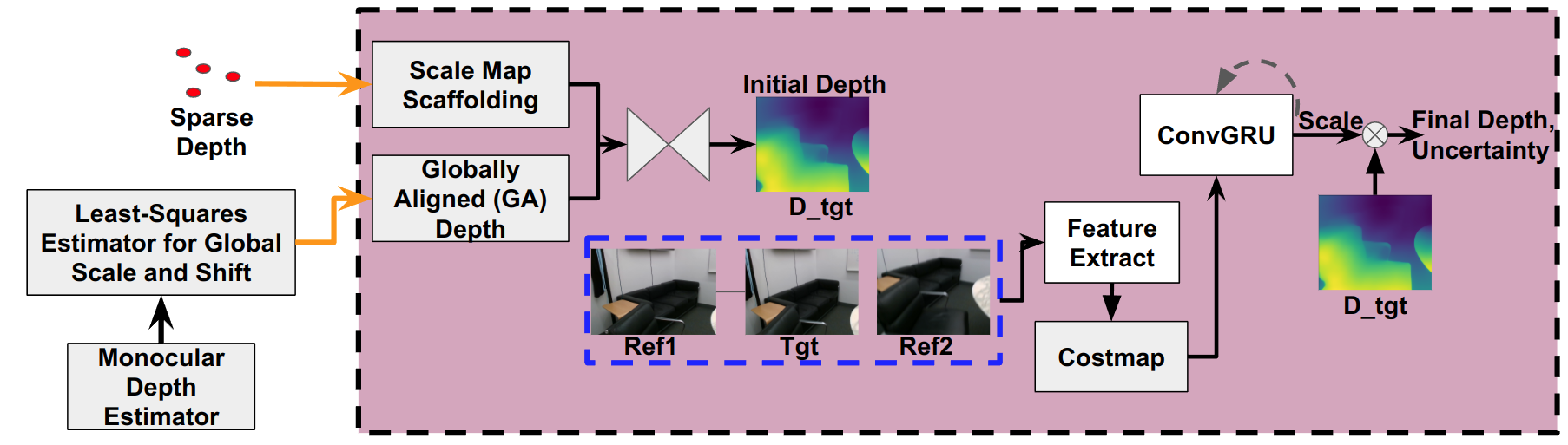}
    \caption{Iterative Refined Metric Depth Module}
    \label{fig:pipeline_2}
  \end{subfigure}
  \caption{The proposed visual-inertial motion and depth (VIMD) learning pipeline.
  (\subref{fig:pipeline_1}) System overview: 
  The VIO filter efficiently fuses RGB images and IMU data to estimate sparse features' metric depth and camera poses, which are then passed to the iterative depth refinement module to predict dense metric depth and its uncertainty. 
  (\subref{fig:pipeline_2}) Iterative refined metric depth module:
  The initial metric depth is estimated using the global alignment (GA) depth, which is the metric-aligned depth from the monocular depth estimator, fitted with a global scale and offset via least squares using the sparse depth. 
  Reference frames are warped to the target frame using the predicted depth and VIO poses, and the scale is iteratively refined using a ConvGRU to predict the final depth and uncertainty. 
  Multi-view information is leveraged to improve the accuracy and robustness of depth estimation.}
  \label{fig:pipeline_overview}
\end{figure*}
% \todo{update the fig: vio->motion estimation}

% \subsection{System overview}

We especially target metrically accurate and robust dense depth in monocular visual–inertial settings
and develop an efficient modular visual–inertial motion and depth (VIMD) learning algorithm as depicted in Fig.~\ref{fig:pipeline_overview}.
Unlike latent-code methods such as CodeVIO~\cite{Zuo2021ICRA}, DiT-SLAM~\cite{Zhao2022_MDPI}, and AB-VINS~\cite{Merrill2024_arxiv}, we do not optimize a compact depth code inside the estimator, avoiding the need for online code optimization and restrictive parameterizations. Instead, we recover per-pixel metric depth by formulating sparse-to-dense alignment as a learning-based optimization problem guided by multi-view geometric constraints from VIO poses. 
Compared to VI-Depth~\cite{Wofk2023_ICRA} and VOICED~\cite{Wong2020_RAL}, the proposed VIMD fully exploits temporal consistency and multi-view geometry, improving robustness in low-density regions and yielding more reliable depth alongside uncertainty estimates.

In particular, 
for simplicity, the proposed VIMD pipeline assumes synchronized IMU measurements together with RGB images to estimate camera poses, which however can be easily relaxed by performing online calibration~\cite{Yang2023TRO}.
The inclusion of IMU data in principle enables recovery of the metric scale of monocular vision. 
Given an image sequence and synchronized IMU measurements, we employ an efficient MSCKF-based~\cite{Mourikis2007_ICRA} VIO module to compute the camera motion and obtain a set of sparse 3D metric points.
These metric points are used both to compute a global scale and bias (offset) for depth alignment and to generate a scale–map scaffold for the depth network. 
The estimated camera poses are also used to enforce the multi-view constraints, thus improving the robustness and accuracy of depth learning.

As the monocular motion estimation module largely leverages the prior VIO work~\cite{Geneva2020ICRA}, which is briefly described in Appendix~\ref{sec:vio},
in the following, we present in detail the design of {\em Iterative Refined Metric Depth Module} (see Fig.~\ref{fig:pipeline_2}), 
serving as the key contribution of this paper.

\subsection{Global Alignment}

Leveraging the MiDaS DPT-Hybrid network~\cite{Ranftl2021_ICCV}, which produces ``affine-invariant'' inverse depth maps $D_{\mathrm{inv}}(u,v)$, for all pixels $(u,v)$, with the corresponding raw depth map 
% \begin{align}
$D(u,v) = \frac{1}{D_{\mathrm{inv}}(u,v)}$,
% \end{align}
we recover the metric scale by a least-squares alignment between the predicted inverse depth and the sparse features' metric depths $1/||{}^{C}\mathbf{p}_{f_i}||$ available from the VIO module (see Appendix~\ref{sec:vio}).
The aligned inverse depth is modeled with the following affine relationship:
\begin{align}
    Z_{\mathrm{inv}}^{\mathrm{GA}}(u,v) &= s_{\mathrm{inv}}\,D_{\mathrm{inv}}(u,v) + b_{\mathrm{inv}}
\end{align}
where $s_{\mathrm{inv}}$ and $b_{\mathrm{inv}}$ are the global scale and offset, respectively, and obtained from the least-squares fitting.

\subsection{Scale-Map Scaffold}

While the previous global alignment corrects the average scale, local residuals may remain inconsistent.
Inspired by~\cite{Wofk2023_ICRA}, we construct a spatially varying \emph{scale map scaffold} from the sparse metric points. 
Let $\mathcal{K}=\{(x_j,y_j)\}_{j=1}^{N_k}$ denote the image coordinates of valid sparse depth samples. 
At each knot, we compute a local scale factor by comparing the sparse metric inverse depth $Z_{\mathrm{inv}}(x_j,y_j)$ with the network prediction $D_{\mathrm{inv}}(x_j,y_j)$:
% \begin{align}
    $\sigma_j = \frac{Z_{\mathrm{inv}}^{\mathrm{GA}}(x_j,y_j)}{D_{\mathrm{inv}}(x_j,y_j)}$.
% \end{align}
The dense scale map $S(u,v)$ is then obtained by interpolating these knot values across the image:
\begin{align}
    S(u,v) = \mathrm{Interp}\!\left(\mathcal{K}, \{\sigma_j\}, (u,v), \text{linear}\right)
\end{align}
where $\mathrm{Interp}(\cdot)$ denotes 2D scattered linear interpolation.  
This map serves as a per-pixel prior, normalized to unit range, and is provided as an additional input to the refinement network to guide predictions toward metric depth.

\subsection{Iterative Refinement}

Given the globally aligned depth and the scaffold, the refinement network predicts an initial metric depth which is then iteratively corrected using multi-view temporal consistency.  
Specifically, as shown in Fig.~\ref{fig:pipeline_2}, we refine depth using three frames: one target $I_t$ and two references $\{I_r\}$, with camera intrinsics $\mathbf K$ and known VIO poses $\{{}_{G}^{I_k}\bar{q}, {}^{G}\mathbf{p}_{I_k}\}$.  
Each image is resized to $224{\times}384$ for the DPT-Hybrid backbone, which outputs predictions at this resolution.  
The depth maps are bicubically upsampled to the native resolution $H\times W$, where supervision is applied against full-resolution ground truth, as 
we found that full-resolution supervision improves scale estimation (see Tab.~\ref{tab:ablations}).
The main steps of this refinement are the  following:

\subsubsection{Feature extraction}  
A lightweight ResNet-18 backbone with stride 4 extracts features $\mathbf{F}_t,\mathbf{F}_r \in \mathbb{R}^{B\times64\times H_c\times W_c}$, where $(H_c,W_c)=(H/4,W/4)$. 
Then a context encoder processes the globally aligned inverse depth $Z_{\mathrm{inv}}^{\mathrm{GA}}$ and the dense scaffold $S(u,v)$, producing a 160-dimensional feature map, which is subsequently divided to form the initial hidden state $\mathbf{h}^{(0)} \in \mathbb{R}^{B\times128\times H_c\times W_c}$ and context features $\mathbf{f}_{\text{context}} \in \mathbb{R}^{B\times32\times H_c\times W_c}$.
A small convolutional head predicts an initial per-pixel scale $\Delta s_{\mathrm{init}}$ and log-variance $\log\sigma^2_{\mathrm{init}}$, yielding the starting inverse depth:
\begin{align}
    Z_{\mathrm{inv}}^{(0)} = Z_{\mathrm{inv}}^{\mathrm{GA}} \odot \Delta s_{\mathrm{init}}
\end{align}

\subsubsection{Cost computation}
At iteration $k$, geometric consistency is enforced by warping each reference feature map into the target view using the current depth:
\begin{align}
    \tilde{\mathbf{F}}_{r_i\rightarrow t} 
    = \mathrm{warp}\!\left(\mathbf{F}_{r_i}, Z_{\mathrm{inv}}^{(k)}, {}^{G}\mathbf{p}_{I_t}, {}^{G}\mathbf{p}_{I_{r_i}}, \mathbf K\right)
\end{align}
Cosine similarity with validity masking yields per-reference costs which are then averaged across all references:
\begin{align}
    \mathbf{c}_{r_i} &= \Big(1 - \langle \mathrm{norm}(\mathbf{F}_t), \mathrm{norm}(\tilde{\mathbf{F}}_{r_i\rightarrow t}) \rangle \Big) \odot \mathbf{M}_{r_i} \\
 \Rightarrow~   \mathbf{c} &= \tfrac{1}{R}\sum_{i=1}^R \mathbf{c}_{r_i}
\end{align}

\subsubsection{Projection and recurrent update}  
To combine the cost with the current depth hypothesis, we use a projection layer that processes both $\mathbf{c}$ and $Z_{\mathrm{inv}}^{(k)}$ with separate convolutional streams and merges them into a joint representation:
\begin{align}
    \mathbf{p}^{(k)} = \mathrm{Proj} \big(Z_{\mathrm{inv}}^{(k)}, \mathbf{c}\big)\in\mathbb{R}^{B\times32\times H_c\times W_c}
\end{align}
The concatenation of $\mathbf{p}^{(k)}$ and $\mathbf{f}_{\text{context}}$ is fed to a separable ConvGRU with hidden dimension 128.  
Following~\cite{Teed2020_ECCV, Gu2023_RAL}, gated updates are applied first horizontally ($1{\times}5$) and then vertically ($5{\times}1$), yielding
\begin{align}
    \mathbf{h}^{(k+1)} = \mathrm{StepConvGRU}\!\big(\mathbf{h}^{(k)}, [\mathbf{p}^{(k)}, \mathbf{f}_{\text{context}}]\big)
\end{align}
from which the network predicts a multiplicative scale correction and log-variance:
\begin{align}
    \Delta s^{(k)} &= 1 + 0.5 \tanh(\phi(\mathbf{h}^{(k+1)})) \\
    \log\sigma^{2(k)} &= \psi(\mathbf{h}^{(k+1)})
\end{align}
The inverse depth is then updated and clamped:
\begin{align}
    Z_{\mathrm{inv}}^{(k+1)} = \mathrm{clamp}\!\left(Z_{\mathrm{inv}}^{(k)} \odot \Delta s^{(k)}, \tfrac{1}{Z_{\max}}, \tfrac{1}{Z_{\min}}\right)
\end{align}

\subsubsection{Final output}  
This refinement is repeated for 3 
%$K_{\mathrm{iter}}{=}3$ 
steps, 
with the cost recomputed at each iteration.  
The final inverse depth is bicubically upsampled to the native resolution $H\times W$ and output along with the corresponding uncertainty map.

\subsection{Loss Function}

Training supervision uses metric ground-truth inverse depth $Z_{\mathrm{inv}}^{\ast}$ at multiple scales.  
At scale $i$, the predicted inverse depth $Z_{\mathrm{inv}}^{[i]}$ is converted to depth $D^{[i]} = 1/Z_{\mathrm{inv}}^{[i]}$ and compared to ground truth $D^{\ast} = 1/Z_{\mathrm{inv}}^{\ast}$ using an uncertainty-weighted Laplace negative log-likelihood:
\begin{align}
    \mathcal{L}^{[i]} &= 
    \frac{|D^{[i]} - D^{\ast}|}{b^{[i]}} + \log b^{[i]} \\
    \quad b^{[i]} &= \exp\!\big(\log\sigma^{2[i]}\big) + \epsilon
\end{align}
Pixels outside the valid depth range $[D_{\min},D_{\max}]$ are masked out. 
In our tests, we set $D_{\min} = 0.1 m$  and $D_{\max} = 5 m$.
The total loss is an exponentially weighted average across all $s$ scales:
\begin{align}
    \mathcal{L}_{\text{depth}} \;=\; 
    \frac{\sum_{i=1}^{s} \gamma^{s-i}\,\mathcal{L}^{[i]}}{\sum_{i=1}^{s} \gamma^{s-i}}
\end{align}
where $\gamma\in(0,1)$ controls the relative contribution of coarse and fine scales (for example, we use $\gamma=0.85$).  
This formulation encourages both accurate depth prediction and per-pixel uncertainty.

\section{Experimental Results}
\label{sec:exp}

% \subsection{Setup and metrics}

Our monocular visual-inertial motion estimation module is built on top of OpenVINS~\cite{Geneva2020ICRA} 
(also see Appendix~\ref{sec:vio}),
while MiDaS DPT Hybrid \cite{Ranftl2021_ICCV} is used as the depth estimator and ResNet-18 as the backbone network initialized with pretrained ImageNet \cite{Deng2009_CVPR} weights, while the remaining layers are initialized randomly. 
We employ the AdamW optimizer \cite{Loshchilov2017_ICLR} with $\beta_1 = 0.9$, $\beta_2 = 0.999$, and weight decay $\lambda = 0.001$. The learning rate is set to $10^{-4}$ for the VOID dataset \cite{Wong2020_RAL} and $10^{-5}$ for the TartanAir \cite{Wang2020_IROS}. 
A step-based scheduler is used, which halves the learning rate every 8 epochs. Training is performed for 25 epochs on a node with $2\times$ NVIDIA A4500 GPUs, using a batch size of 8. The entire training process takes approximately 3 days.

We follow the evaluation protocol~\cite{Wofk2023_ICRA} and consider ground truth depth to be valid between 0.2 and 5 meters. 
The minimum and maximum depth prediction values in these works are set to 0.1 and 8.0 meters, respectively. 
We clamp depth predictions, both after global alignment and after applying regressed dense scale maps, to this range.
We evaluate in inverse depth space $z = 1/D(u,v)$ (in $\mathrm{km}^{-1}$), which penalizes errors at closer ranges more,
with the following metrics:
$\mathrm{iMAE} = \frac{1}{M} \sum_{i} | z_i^{gt} - \hat{z}_i |$, \ 
$\mathrm{iRMSE} = \sqrt{\frac{1}{M} \sum_{i} ( z_i^{gt} - \hat{z}_i )^2}$, \ 
$\mathrm{iAbsRel} = \frac{1}{M} \sum_{i} \frac{| z_i^{gt} - \hat{z}_i |}{z_i^{gt}}$,
and also compute MAE and RMSE in  depth space $D(u,v)$ (in mm).

\begin{table}%[ht]
\centering
\setlength{\tabcolsep}{9pt} % Reduce column separation
\begin{tabular}{lrrrr}
\hline
\textbf{Method} & \textbf{RMSE} & \textbf{MAE} & \textbf{iRMSE} & \textbf{iMAE} \\
\hline
GA Only    & 249.40        & 169.45       & 106.17        & 74.57        \\
VOICED-S$^*$ \cite{Wong2020_RAL}   & 253.14        & 131.54       & 126.30        & 87.36        \\
KBNet$^*$ \cite{Wong2021_ICCV}      & 263.54        & 131.54       & 128.29        & 66.84        \\
SML$^*$ \cite{Wofk2023_ICRA} & 167.15 & 97.03 & 74.67 & 46.62 \\
\textbf{Ours} & \textbf{155.23} & \textbf{88.42} & \textbf{67.21} & \textbf{40.18} \\
\hline
\end{tabular}
\caption{
Quantitative results of evaluation on VOID. 
All the methods use DPT-H as the depth model and 150 sparse depth points. %Lower is better. \\
$^*$Results are taken from the original papers.
}
\label{tab:void_150_tab}
\end{table}

\begin{table*} %[ht]
\centering
\small
\begin{tabular}{c
                r r r
                @{\hskip 10pt} r r r
                @{\hskip 10pt} r r r
                @{\hskip 10pt} r r r}
\toprule
\multirow{2}{*}{\textbf{Reduction \%}} &
\multicolumn{3}{c}{\textbf{RMSE}} &
\multicolumn{3}{c}{\textbf{MAE}} &
\multicolumn{3}{c}{\textbf{iRMSE}} &
\multicolumn{3}{c}{\textbf{iMAE}} \\
\cmidrule(lr){2-4} \cmidrule(lr){5-7} \cmidrule(lr){8-10} \cmidrule(lr){11-13}
 & \textbf{GA} & \textbf{SML} & \textbf{Ours} 
 & \textbf{GA} & \textbf{SML} & \textbf{Ours} 
 & \textbf{GA} & \textbf{SML} & \textbf{Ours} 
 & \textbf{GA} & \textbf{SML} & \textbf{Ours} \\
\midrule
10  & 249.65 & 169.23 & \textbf{158.74} & 169.51 & 97.03  & \textbf{89.52}  & 106.27 & 75.62  & \textbf{69.19}  & 74.61  & 45.09  & \textbf{41.08}  \\
50  & 251.71 & 180.26 & \textbf{169.87} & 171.28 & 104.07 & \textbf{96.57}  & 107.00 & 80.24  & \textbf{74.95}  & 75.35  & 48.38  & \textbf{44.57}  \\
80  & 255.53 & 216.55 & \textbf{194.88} & 174.14 & 129.83 & \textbf{112.47} & 108.96 & 98.61  & \textbf{94.47}  & 76.63  & 60.63  & \textbf{53.17}  \\
\bottomrule
\end{tabular}
\caption{Performance with respect to  different levels of sparse depth reduction.}
\label{tab:combined_metrics_selected_reduce}
\end{table*}

%%%%%%%%%%%%%%%%%%%%
\subsection{Evaluation on VOID} \label{sec:void}

\begin{figure}
    \centering
    \includegraphics[width=1.0\linewidth]{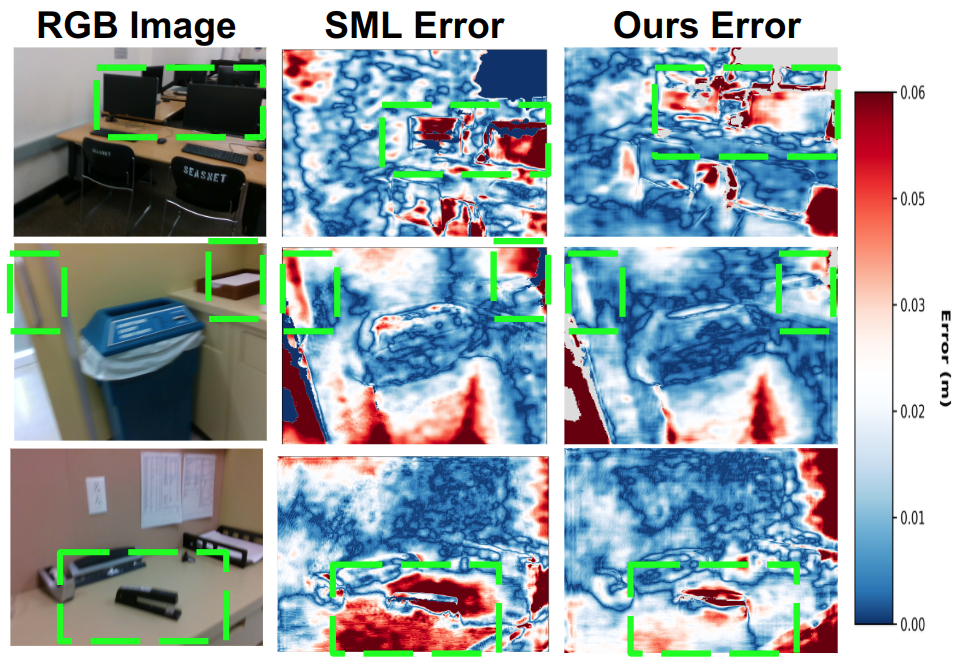}\vspace{5pt}
    \includegraphics[width=1.0\linewidth]{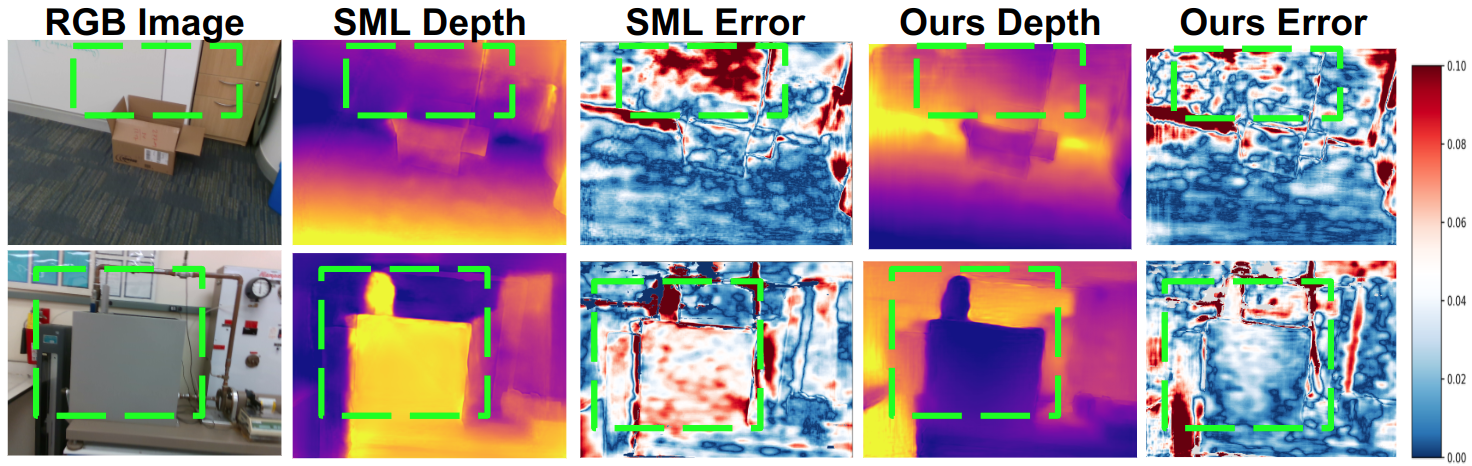}
    \caption{Qualitative results of evaluation on VOID.
     % (150 samples). 
    % Note that in error maps, red is positive depth error (farther from the GT depth) while blue is low error (closer to the GT depth).
    }
    \label{fig:void_depth_viz1}
\end{figure}

% \begin{figure}
%     \centering
%     \includegraphics[width=1.0\linewidth]{figures/void_res_viz2.png}
%     \caption{Our method tested on VOID 150 samples. In error maps, red is positive depth error (farther from the GT depth) and blue is low error (closer to the GT depth).}
%     \label{fig:void_res_viz2}
% \end{figure}

For fair comparison with SOTA methods, we first evaluate our method on the VOID dataset \cite{Wong2020_RAL}, which provides real-world data collected using the Intel RealSense D435i camera and a VIO system. 
We use 150 samples, consistent with the typical number of observations produced by a lightweight VIO system (120–225 point features under this dataset). 
We particularly focus on the results against the SML~\cite{Wofk2023_ICRA}, as it was trained on VOID under the same setting.
We note that the distribution of sparse points has a significant impact on the network’s performance. 
Since the point distribution generated by our VIO differs from that of VOID, we ensure fairness by training and evaluating both the SML and our method directly on VOID.

Tab. \ref{tab:void_150_tab} shows the quantitative results.
Our method achieves iRMSE and iMAE improvements of 37\% and 46\% respectively with the GA depth, corresponding to relative error reductions of 10\% (iRMSE) and 14\% (iMAE) compared to the SML.
Fig. \ref{fig:void_depth_viz1} depicts the qualitative results of depth predictions.
Note that error maps are computed against the ground truth depth, where blue indicates low error and red indicates high error. 
It is clear that as compared to the SML, our method reduces errors particularly in central regions of the depth maps, through iterative scale refinement.
%
% Further examples are shown in Fig. \ref{fig:void_res_viz2}. 
Importantly, our predictions are not only metrically more accurate but also produce sharper object boundaries.
For instance, in the bottom row of Fig. \ref{fig:void_depth_viz1}, the vertical pole is reconstructed with reduced error compared to the SML, highlighting the accuracy of our refinement strategy.

More interestingly, beyond absolute accuracy, Tab.~\ref{tab:combined_metrics_selected}  shows the comparative results under different levels of sparse point reduction. Even when the number of points is reduced by 80\%, our method consistently outperforms GA and SML across all metrics. Notably, the relative advantage over SML increases as the point density decreases, highlighting the robustness of our approach.
This robustness arises from two key factors. First, our method leverages multi-view information through pose integration, which provides additional geometric constraints that compensate for the reduced spatial coverage of sparse points. Second, the ConvGRU module learns to iteratively refine the scale, enabling the network to progressively optimize depth estimates across frames. By combining multi-view cues with recurrent scale refinement, our model is able to maintain both accuracy and stability, even during low density sparse points.

\begin{figure}
    \centering
    \includegraphics[width=1.0\linewidth]{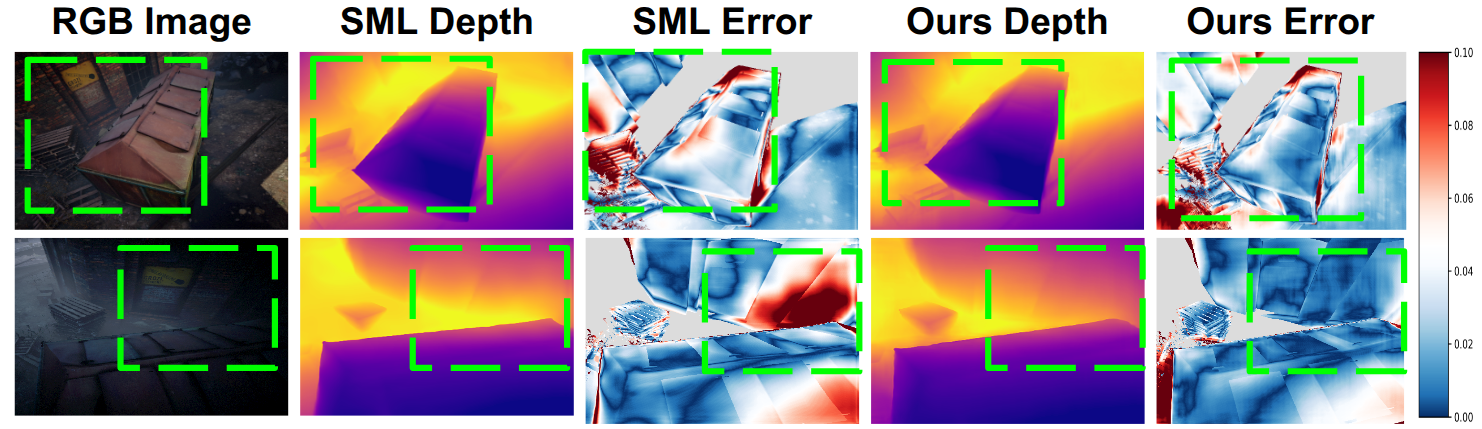}
    \caption{Qualitative results of evaluation on TartanAir.
     % (150 samples). 
    % Note that in error maps, red is positive depth error (farther from the GT depth) while blue is low error (closer to the GT depth).
    }
    \label{fig:tartan_depth_viz1}
\end{figure}

\subsection{Evaluation on TartanAir} \label{sec:tartanair}

% There is not a lot of large scale training data (RGBD+IMU) for training. We use TartanAir \cite{Wang2020_IROS} dataset. We use a 70-30 train-test split samples are taken from both easy and hard seqeunces. We cleaned the data if there are any bad runs and bad depth due to frontend. We run openVINS \cite{Geneva2020ICRA} ov_sim package on tartan dataset by using splines and syncing images and collected sparse depth (MSCKF + SLAM points) and camera poses. The sparse depth distribution is different from the VOID dataset and \cite{Wofk2023_ICRA} fails on the collected dataset hence we re-train and compare with our pipeline.

To further evaluate the proposed method in large-scale scenarios
(where real training data of synchronized RGB-D and IMU is limited),
we have leveraged the TartanAir dataset \cite{Wang2020_IROS}, which offers photo-realistic RGB-D sequences generated in diverse simulated environments. 
We adopt a 70--30 train/test split, sampling sequences from both \textit{easy} and \textit{hard} difficulty levels. Prior to training, we cleaned the data by removing runs with invalid trajectories or corrupted depth maps caused by failures in the rendering front-end.  
Since the TartanAir does not provide IMU measurements, we generated inertial data by fitting splines (using the \texttt{ov\_sim} package \cite{Geneva2020ICRA}) to the ground-truth poses and differentiating them to obtain accelerations and angular velocities at 200~Hz. The simulated IMU follows realistic noise characteristics, with an accelerometer noise density of $2.0 \times 10^{-3}$~m/s$^2$/\(\sqrt{\text{Hz}}\), an accelerometer random walk of $3.0 \times 10^{-3}$~m/s$^3$/\(\sqrt{\text{Hz}}\), a gyroscope noise density of $1.6968 \times 10^{-4}$~rad/s/\(\sqrt{\text{Hz}}\), and a gyroscope random walk of $1.9393 \times 10^{-5}$~rad/s$^2$/\(\sqrt{\text{Hz}}\).  
Using these simulated IMU signals, we ran the OpenVINS \cite{Geneva2020ICRA} on the TartanAir sequences. 
Spline interpolation was applied to synchronize visual and inertial streams, and we collected sparse depth points (from MSCKF and SLAM features) along with camera poses. This resulted in a dataset of images, sparse depths, poses that closely reflect the outputs of a real-world visual-inertial motion estimation system.  
The distribution of sparse depth points in TartanAir differs from that of VOID. 
Consequently, the pretrained network from \cite{Wofk2023_ICRA} fails to generalize when applied directly to our collected data. 
For a fair comparison, we implemented the training pipeline for the baseline method and retrained it on the processed TartanAir dataset alongside the proposed approach.

Fig. \ref{fig:tartan_depth_viz1} shows qualitative results and Tab.~\ref{tab:tartanair_tab} reports the quantitative results on TartanAir. 
As evident, our method achieves the lowest error across all metrics, outperforming the SML \cite{Wofk2023_ICRA} by 4.3\% in RMSE, 5.8\% in MAE, 6.9\% in iRMSE, and 8.8\% in iMAE. 
Beyond accuracy, our model is more compact, with a size of {15.1~MB} compared to {21.3~MB} for the SML. 
In terms of runtime, our method achieves 19.6\,ms per frame on an NVIDIA RTX~2080Ti GPU, while the SML runs slightly faster at 15.3\,ms. 
This additional overhead primarily comes from the GRU refinement module, which can be further optimized through TensorRT \cite{tensorrt}.
%The GA-only baseline, which does not perform learned depth refinement, performs significantly worse across all metrics, underscoring the importance of our refinement strategy. 
%
%depth infer: 54ms scaffolding 16.2ms 
%
These results highlight the benefit of incorporating pose-based multi-view constraints and iterative scale refinement into the depth prediction pipeline. 
By leveraging both sparse points and temporal consistency, our method not only improves predictive accuracy but reduces model complexity, making it more practical for deployment in resource-constrained settings.  

% This effort demonstrates the additional steps we undertook to enable rigorous evaluation: beyond simply using an existing dataset, we generated realistic IMU data, integrated it into the OpenVINS pipeline to produce sparse depth and poses, and built a complete training framework for both our method and the baseline. This ensures a meaningful and controlled evaluation on large-scale RGB-D+IMU data.    

\begin{table}
\centering
\setlength{\tabcolsep}{4pt} % Reduce column separation
\begin{tabular}{lrrrrrr}
\hline
\textbf{Method}  & \textbf{RMSE} & \textbf{MAE} & \textbf{iRMSE} & \textbf{iMAE} & \textbf{Size(MB)} & \textbf{Time(ms)} \\
\hline
GA Only      & 231.86 & 231.86 & 31.68 & 21.81 & --- & ---\\
SML \cite{Wofk2023_ICRA} & 180.53 & 110.33 & 27.61 & 17.24 & 21.3 & \textbf{15.3} \\ 
Ours        & \textbf{172.81} & \textbf{103.92} & \textbf{25.70} & \textbf{15.72} & \textbf{15.1} & 19.6 \\
\hline
\end{tabular}
\caption{Qualitative results of evaluation on TartanAir v1. 
% All methods use DPT-H as the depth model and 70--150 sparse depth points. Lower is better.
}
\label{tab:tartanair_tab}
\end{table}

% \begin{table}[ht]
% \centering
% \begin{tabular}{lrrrr}
% \hline
% \textbf{Method} & \textbf{RMSE} & \textbf{MAE} & \textbf{iRMSE} & \textbf{iMAE} \\
% \hline
% GA Only       \\
% SML \cite{Wofk2023_ICRA}  \\ 
% Ours        \\
% \end{tabular}
% \caption{
% Zero shot evaluation on Euroc Mav. trained on the tartanV1. All methods use DPT-H as the depth model and 70-150 sparse depth points. Lower is better.}
% \end{table}

% While full-resolution supervision is common in standard depth nets [Godard19], we show it is critical for networks that learn a per-pixel scale map, yielding a further 4 \% RMSE drop (Table 3).

\subsection{Evaluation on AR Table}

To evaluate the generalization ability of our model, we trained exclusively on TartanAir and then performed {\em zero-shot} deployment on the AR Table dataset \cite{Chen2023ICRA}. The AR Table sequences consist of real-world RGB-D+IMU data stored in rosbags. 
For evaluation, we used a minute of data and filtered out outlier sparse depths with errors greater than 8\,cm. Quantitative results are summarized in Tab.~\ref{tab:table_zeroshot}.  
Across both Table~1 and Table~2 sequences, our method consistently outperforms the GA-only baseline and the SML approach. Without iterative refinement (0 iter), our method already achieves competitive accuracy, reducing both RMSE and MAE relative to SML. When iterative refinement is enabled (3 iters), errors decrease further: on Table~1, RMSE improves from 547.63 to 541.02, while AbsRel is reduced from 0.120 to 0.117. On Table~2, RMSE decreases from 368.10 to 357.47 and AbsRel from 0.131 to 0.111. 
These results highlight that iterative scale and geometry refinement aids to further reduce the depth error.    

Importantly, this zero-shot experiment demonstrates that our method generalizes effectively to real-world datasets despite training in simulated drone environments (TartanAir). 
The improvements obtained through iterative refinement indicate that the model is capable of leveraging underlying geometric cues at test time, which is crucial for practical deployment in diverse environments where the training and deployment domains may differ significantly.

\begin{table} %[ht]
\centering
\setlength{\tabcolsep}{10pt} % Reduce column separation
\begin{tabular}{l l rrr}
\hline
\textbf{Dataset} & \textbf{Method} & \textbf{RMSE} & \textbf{MAE} & \textbf{AbsRel} \\
\hline
\multirow{3}{*}{Table 1} & GA   &657.23&  459.52&   0.391   \\
                         & SML   &563.47   &334.84   &0.121   \\
                         & Ours (0 iter) &547.63 &330.10 &0.120 \\
                         & Ours (3 iter) &\textbf{541.02}&   \textbf{326.97}&   \textbf{0.117}   \\
                         
\hline
\multirow{3}{*}{Table 2} & GA    &626.71&   251.98&   0.140   \\
                         & SML   &387.08&   186.43&   0.116   \\
                         & Ours (0 iter) &368.10 &192.34 &0.131 \\
                         & Ours (3 iter) &\textbf{357.47}&   \textbf{185.54}&   \textbf{0.111}   \\
% \hline
% \multirow{3}{*}{Table 3} & GA & 743.27  & 333.56  & 0.160  \\
%                          & SML   &   &   &   \\
%                          & Ours  &411.50   &240.04   &0.113    \\
% \hline
% \multirow{3}{*}{Table 4} & GA & 70.30  &55.31   &0.143   \\
%                          & SML   &   &   &   \\
%                          & Ours  &40.43   &28.34   &0.073   \\
% \hline
% \multirow{3}{*}{Table 5} & GA &76.95   &55.65   &0.163   \\
%                          & SML   &   &   &   \\
%                          & Ours  &60.54   &36.81   &0.107   \\
% \hline
% \multirow{3}{*}{Table 6} & GA &93.57   &67.49   &0.171   \\
%                          & SML   &85.18   &58.22   &0.150   \\
%                          & Ours  &74.20   &44.45   &0.114   \\
% \hline
% \multirow{3}{*}{Table 7} & GA &87.03   &65.22   &0.177   \\
%                          & SML   &   &   &   \\
%                          & Ours  &68.92   &43.98   &0.118   \\
% \hline
% \multirow{3}{*}{Table 8} & GA &38.21   &27.37   &0.081   \\
%                          & SML   &   &   &   \\
%                          & Ours  &28.10   &16.35   &0.049   \\
% \hline
% \multirow{3}{*}{Average} & GA &   &   &   \\
%                          & SML   &   &   &   \\
%                          & Ours  &   &   &   \\
\hline
\end{tabular}
\caption{Evaluation results on the AR Table datasets.}
\label{tab:table_zeroshot}
\end{table}

%%%%%%%%%%%%%%%
\begin{table*} %[t]
\centering
\small
\setlength{\tabcolsep}{15pt} % Reduce column separation
\newcommand{\cmark}{\ding{51}}  
\newcommand{\xmark}{\ding{55}}  
\begin{tabular}{lccccccc}
\toprule
\multirow{2}{*}{Loss Used} &
\multicolumn{4}{c}{Lower is better $\downarrow$} & 
\multicolumn{3}{c}{Ablation switches} \\
\cmidrule(lr){2-5}\cmidrule(l){6-8}
& RMSE & MAE & iRMSE & iMAE & \textbf{Full-res} & \textbf{GRU Refinement} & \textbf{Uncert.} \\
\midrule
L1                 & \textit{167.15} & \textit{95.69} & \textit{75.42} & \textit{44.69} & \xmark & \xmark & \xmark \\
L1                     & \textit{163.70} & \textit{91.18} & \textit{71.14} & \textit{41.17} & \cmark & \xmark & \xmark \\
L1             & \textit{155.23} & \textit{88.42} & \textit{67.21} & \textit{40.18} & \cmark & \cmark & \xmark \\
Laplace       & \textit{158.41} & \textit{86.72} & \textit{66.48} & \textit{38.22} & \cmark & \cmark & \cmark \\
Gaussian  & \textit{-} & \textit{-} & \textit{-} & \textit{-} & \cmark & \xmark & \cmark \\
\bottomrule
\end{tabular}
\caption{Ablation study of the proposed VIMD on the VOID-150 dataset.  
“Full-res” means the depth prediction (144\,$\times$\,284) is bilinearly up-sampled to 480\,$\times$\,640 and supervised at native resolution.  
“Warp” enables ConvGRU multi-view warping.  
“Uncert.” indicates the Laplace negative-log-likelihood head is used instead of pure L1.  
% Replace italic placeholders with your measured scores.
}
\label{tab:ablations}
\end{table*}
%%%%%%%%%%

\subsection{Ablation Study}

We have further conducted an ablation study on the VOID-150 dataset to understand the contribution of different components in the proposed VIMD model. 
Tab.~\ref{tab:ablations} summarizes the results.  
The first factor examined is prediction resolution. Training with only low-resolution L1 supervision produces the largest error, consistent with prior findings in \cite{Wofk2023_ICRA}. This outcome suggests that restricting supervision to a coarse scale limits the network’s ability to recover fine structural details. When full-resolution supervision is introduced by bilinearly up-sampling predictions before supervision, the error decreases noticeably. Our conjecture is that the additional pixel-level guidance forces the network to preserve object boundaries and local depth gradients, which are otherwise smoothed out at lower resolutions.  

We then study the effect of recurrent refinement with ConvGRU. Adding this module consistently improves all metrics, with the most significant gains in iRMSE and iMAE. Unlike static single-pass prediction, the recurrent updates allow the network to iteratively align depth estimates across frames while correcting the global scale. We hypothesize that this iterative mechanism encourages the network to exploit multi-view consistency: features that remain stable across frames are reinforced, while transient ambiguities are suppressed. This explains why improvements are particularly strong in inverse-error metrics, which are sensitive to structural misalignment at different depth ranges.  

We finally investigate uncertainty modeling. Replacing the L1 loss with a Laplace negative log-likelihood not only lowers the overall error but also sharpens predictions in ambiguous regions. We attribute this to the model’s ability to down-weight supervision in areas where sparse depth samples are noisy or inconsistent, allowing it to focus learning on more reliable cues. In contrast, using a Gaussian likelihood did not converge reliably—likely due to the heavier tails of depth errors in real-world sparse inputs, which are better captured by a Laplace distribution.  

Together, these results show that each component tackles a distinct failure mode: full-resolution supervision reduces over-smoothing, the ConvGRU refinement mitigates scale drift while enforcing temporal consistency, and uncertainty modeling enhances robustness to noisy supervision. The combination of all three produces the most accurate and stable predictions compared to other state-of-the-art methods.

\section{Conclusions and Future Work}
\label{sec:concl}

In this paper, we presented VIMD, a monocular visual--inertial motion and depth framework that recovers dense metric depth using only a monocular camera and an IMU. 
At the core of our approach is to leverage multi-view information within a learning-based framework to iteratively optimize per-pixel scale. 
Additionally, uncertainty prediction improves the reliability of the estimated depth for downstream tasks, while full-resolution supervision further enhances final depth accuracy.
%
% As a result, the proposed VIMD model is robust in scenarios with sparse features and demonstrates generalization ability, including zero-shot performance on real-world datasets. 
%
As a result, the VIMD is robust under fewer sparse features and generalizes well across environments, including zero-shot performance on real-world datasets.

We also identify current limitations: iterative refinement can degrade in textureless regions and under inaccurate pose estimates. As a practical mitigation, the proposed refinement module can be applied after sparse bundle adjustment, where more reliable camera poses are available. In the future work, we will extend this model into a tightly-coupled SLAM  by training in an end-to-end fashion. 
Finally, VIMD can serve as a strong metric depth prior for downstream mapping systems, such as Gaussian-Splatting (GS)-SLAM \cite{Katragadda2025ICCV} for monocular language photorealistic mapping.
%

% Note that the proposed VIMD model can be used as a prior to Gaussian-Splatting (GS)-SLAM \cite{Katragadda2025ICCV} for monocular language photorealistic mapping. 

% We also plan to further improve efficiency to enable deployment on resource-constrained  mobile robots and XR devices.

\appendices
\section{Visual-Inertial Motion Estimation}
\label{sec:vio}

The visual-inertial motion estimation module in the proposed VIMD pipeline (see Fig.~\ref{fig:pipeline_overview}) is built on top of the monocular MSCKF-VIO (i.e., OpenVINS~\cite{Geneva2020ICRA}), 
which offers accurate and efficient ego-motion estimates by fusing IMU data and visual measurements of sparse point features
and  is briefly described in the following.
In particular, 
at time $t_{k}$, the system state $\mathbf{x}_k$ consists of the current navigation states $\mathbf{x}_{I_k}$, historical IMU pose clones $\mathbf{x}_{C}$, and a subset of 3D environmental (SLAM) point features, $\mathbf{x}_f$:
\begin{align}
    \mathbf{x}_k
    &= 
    \label{eq:state}
    \begin{bmatrix}
    \mathbf{x}_{I_k}^{\top}
    ~
    \mathbf{x}_{C}^{\top}
    ~
    \mathbf{x}_{f}^{\top}
    % ~
    % \mathbf{x}_\pi^{\top}
    \end{bmatrix}^{\top} 
    \hspace{-1.5mm}
    ,~~
    \mathbf{x}_{C}
    =
    \begin{bmatrix}
    \mathbf{x}_{T_{k}}^{\top} 
    \dots
    \mathbf{x}_{T_{k-c}}^{\top}
    \end{bmatrix}^{\top} 
    \\
    \mathbf{x}_{I_k}
    &=
    \begin{bmatrix}
    \label{eq:state_imu}
    {}_{G}^{I_k}\Bar{q} ^{\top} &
    {}^{G}\mathbf{p}_{I_k} ^{\top} &
    {}^{G}\mathbf{v}_{I_k} ^{\top} &
    \mathbf{b}_{g} ^{\top}&
    \mathbf{b}_{a} ^{\top} 
    \end{bmatrix}^{\top}
    \\
    % %
    \mathbf{x}_f
    & =
    \begin{bmatrix}
    {}^{G}\mathbf{p}_{f_1}^{\top}
    \dots
    {}^{G}\mathbf{p}_{f_g}^{\top}
    \end{bmatrix}^{\top}
\end{align}
where $ {}_{G}^{I}\Bar{q}$ is the unit quaternion (${}_{G}^{I}\mathbf{R}$ in rotation matrix form) that represents the rotation from the global $\{G\}$ to the IMU frame $\{I\}$; 
${}^{G}\mathbf{p}_{I}$, ${}^{G}\mathbf{v}_{I}$, and ${}^{G}\mathbf{p}_{f_i}$ are the IMU position, velocity, and $i$'th point feature position in $\{G\}$;
$\mathbf{b}_{g}$ and $\mathbf{b}_a$ are the gyroscope and accelerometer biases; 
$\mathbf{x}_{T_i}=[{}_{G}^{I_{i}}\Bar{q}^{\top}~ {}^{G}\mathbf{p}_{I_{i}} ^{\top} ]^\top$.
%

%%===============================================================
\subsection{IMU Propagation} \label{sec:propagation}
The inertial kinematics are used to evolve the state from time $t_k$ to $t_{k+1}$ \cite{Chatfield1997,Trawny2005_Q_TR}:
\begin{align}  
\label{eq:imu_eq}
    \mathbf{x}_{I_{k+1}}
    & =
    \mathbf{f}(\mathbf{x}_{I_{k}}, \mathbf{a}_{m_k},
    \boldsymbol{\omega}_{m_k}
    )
\end{align}
where the linear acceleration $\mathbf{a}_{m_k}$ and the angular velocity $\boldsymbol{\omega}_{m_k}$ measurements are contaminated by zero-mean white Gaussian noises.
% Note that all other states are static and have zero dynamics.
% and the non-inerital states such as clones and features have an identity transform since they remain static.
% Features and clones are static thus remain the same after propagation.
% The corresponding covariance matrix propagation can be shown as:
% The discrete-time system noise covariance \mathbf{Q}_k\mathbf{Q}_k and the propagated state covariance can be computed as:
%
% \begin{align}
%     \centering
%     &
%      {\mathbf{P}}_{k+1|k} 
%     = \mathbf{\Phi}_k{\mathbf{P}}_{k|k}
%     \mathbf{\Phi}^{\top}_k + \mathbf{Q}_k 
%     \label{eq:cov_prop}
%     % \\
%     % &
%     % \mathbf{Q}_k
%     % \nonumber
%     % = 
%     % \int_{t_k}^{t_{k+1}}
%     % \mathbf{\Phi}(k+1,\tau)\mathbf{G}(\tau)\mathbf{Q}\mathbf{G}^{\top}(\tau)  \mathbf{\Phi}^{\top}(k+1,\tau)\textrm{d}\tau
% \end{align}
% %
% where \mathbf{P}_{a|b}\mathbf{P}_{a|b} represents the covariance at t_{a}t_{a} computed using all the visual measurements up to timestep t_bt_b; 
% \mathbf{\Phi}_k\mathbf{\Phi}_k is the linearized state transition matrix from time t_kt_k to t_{k+1}t_{k+1}; \mathbf{Q}_k\mathbf{Q}_k is the IMU noise covariance.
%
The MSCKF linearizes this nonlinear model and propagates the state estimate and covariance~\cite{Mourikis2007_ICRA}.

%%===============================================================
\subsection{Visual Update}

The monocular camera provides bearing observations to sparse point features in the environment.
These visual measurements are used to update the motion states with the following measurement function 
(note that we here assume the global 3D feature model~\cite{Geneva2020ICRA}):
\begin{align}
    \mathbf{z}_k
    =
    \label{eq:bearing_mea}
    \mathbf{h}(\mathbf{x}_k)
    +
    \mathbf{n}_k
    &
    =:
    \bm{\Lambda}({}^{C_k}\mathbf{p}_{f})+\mathbf{n}_k
    \\
    {}^{C_k}\mathbf{p}_{f}
    = 
    [x~y~z]^{\top}
    & = 
    {}_{I}^{C}\mathbf{R}{}_{G}^{I_k}
    \mathbf{R}\left(
    {}^{G}\mathbf{p}_{f} - {}^{G}\mathbf{p}_{I_k}
    \right)
    +{}^{C}\mathbf{p}_{I}
    \\
    \bm{\Lambda}([x~y~z]^{\top}) 
    & =: [ x/z ~ y/z ]^{\top}
\end{align}
where $\mathbf{n}_k$ is the white Gaussian bearing measurement noise and $\{{}_{I}^{C}\mathbf{R}, {}^{C}\mathbf{p}_{I}\}$ is the known camera-IMU rigid transformation.
% \footnote{In this paper, we assume the camera intrinsic/extrinsic are calibrated offline for mathematical brevity. However, our system is fully capable of performing online calibration of them allowing better consistency and robustness to the poor calibrations.}
% bearing measurements of the features are tracked which can be related to the state by:
%
Linearizing  \eqref{eq:bearing_mea} yields the following residual:
\begin{align}
    \tilde{\mathbf{z}}_k
    & 
    \simeq 
    \label{eq:lin_bearing_mea}
    \mathbf{H}_k\tilde{\mathbf{x}}_k
    +\mathbf{n}_k
    =
    \mathbf{H}_{T_k}
    \tilde{\mathbf{x}}_{T_k}
    +
    \mathbf{H}_{f_k}
    {}^{G}\tilde{\mathbf{p}}_f
    +
    \mathbf{n}_k
\end{align}
where $\mathbf{H}_{T_k}$ and $\mathbf{H}_{f_k}$ are the measurement Jacobians in respect to the observing pose $\hat{\mathbf{x}}_{T_k}$ and 3D point feature ${}^{G}\hat{\mathbf{p}}_f$.
% \footnote{
% Note that $\hat{\mathbf{x}}$ is used to denote the \textit{current} estimate of a random variable $\mathbf{x}$ with $\tilde{\mathbf{x}} = \mathbf{x} \boxminus \hat{\mathbf{x}}$ denotes the error state.
% For the quaternion error state, we employ JPL multiplicative error \cite{Trawny2005_Q_TR} i.e., $\delta \bar{q} = \bar{q}\otimes \hat{\bar{q}}^{-1} \simeq [
%     \frac{1}{2}\delta\boldsymbol{\theta}^{\top} ~ 1
% ]^{\top}$.
% % The ``$\boxplus$" and ``$\boxminus$"  operations map elements to and from a given manifold and equate to simple ``+" and ``-" for Euclidean vector variables.
% }
We stack the measurements from different times and have:
\begin{align}
    \tilde{\mathbf{z}}_c 
    &=
    \mathbf{H}^{c}_T\tilde{\mathbf{x}}_{C}
    +
    \mathbf{H}^{c}_{f}
    {}^{G}\tilde{\mathbf{p}}_{f}
    +
    \mathbf{n}_c \label{eq:lin_bearing_measurement}
\end{align}
where $\tilde{\mathbf{z}}_c$ is the stacked measurement residual;
$\mathbf{H}^{c}_T$ and $\mathbf{H}^{c}_{f}$ are the stacked Jacobians;
$\mathbf{n}_c \sim\mathcal{N}(\mathbf{0}, \mathbf{R}_c)$ is the stacked measurement noise (normally 1 pixel).
We then perform EKF update with two types of point features: 
(i) {\em SLAM Point:} The state $\mathbf{x}_{f}$ contains ${}^{G}\mathbf{p}_{f}$, thus \eqref{eq:lin_bearing_measurement} can directly update the state using the standard EKF equations.
(ii) {\em MSCKF Point:} For features that are not in the state, we project  \eqref{eq:lin_bearing_measurement} onto the left nullspace of $\mathbf{H}^c_{f}$ (i.e., $\mathbf{N}_{f}^{\top}\mathbf{H}^{c}_{f}=\mathbf{0}$),
and construct the following residual independent of ${}^{G}\tilde{\mathbf{p}}_{f}$ for update~\cite{Mourikis2007_ICRA}:
\begin{align}
    \mathbf{N}_{f}^{\top}\tilde{\mathbf{z}}_c 
    &=
\mathbf{N}_{f}^{\top}\mathbf{H}^{c}_T\tilde{\mathbf{x}}_{C}
    +
    \mathbf{N}_{f}^{\top}\mathbf{H}^{c}_{f}
    {}^{G}\tilde{\mathbf{p}}_{f}
    +
    \mathbf{N}_{f}^{\top}\mathbf{n}_c
    \\
    \Rightarrow~
    \tilde{\mathbf{z}}_c' 
    &= 
    \label{eq:msckf_feat}
    \mathbf{H}'_T\tilde{\mathbf{x}}_{C}
    +
    \mathbf{n}'
\end{align}
{
% \newpage
% \vspace{0.05cm}
% \def\bibfont{\footnotesize}
% \def\bibfont{\scriptsize}
% \printbibliography
\bibliographystyle{packages/IEEEtran}
\bibliography{libraries/extra,libraries/related,libraries/rpng}

@manual{tensorrt,
  title        = {NVIDIA TensorRT},
  organization = {NVIDIA Corporation},
  year         = {2025},
  note         = {High-performance deep learning inference SDK},
  url          = {https://developer.nvidia.com/tensorrt},
}

@book{Chatfield1997,
	title        = {Fundamentals of High Accuracy Inertial Navigation},
	author       = {Averil B. Chatfield},
	year         = 1997,
	publisher    = {American Institute of Aeronautics and Astronautics, Inc.},
	address      = {Reston, VA}
}

@techreport{Trawny2005_Q_TR,
	title        = {Indirect {K}alman Filter for {3D} Attitude Estimation},
	author       = {Nikolas Trawny and Stergios I. Roumeliotis},
	year         = 2005,
	month        = mar,
	institution  = {University of Minnesota, Dept. of Comp. Sci. \& Eng.}
}

@inproceedings{Deng2009_CVPR,
  title={Imagenet: A large-scale hierarchical image database},
  author={Deng, Jia and Dong, Wei and Socher, Richard and Li, Li-Jia and Li, Kai and Fei-Fei, Li},
  booktitle={2009 IEEE conference on computer vision and pattern recognition},
  pages={248--255},
  year={2009},
  organization={Ieee}
}

@inproceedings{Wang2020_IROS,
  title={Tartanair: A dataset to push the limits of visual slam},
  author={Wang, Wenshan and Zhu, Delong and Wang, Xiangwei and Hu, Yaoyu and Qiu, Yuheng and Wang, Chen and Hu, Yafei and Kapoor, Ashish and Scherer, Sebastian},
  booktitle={2020 IEEE/RSJ International Conference on Intelligent Robots and Systems (IROS)},
  pages={4909--4916},
  year={2020},
  organization={IEEE}
}

@inproceedings{Loshchilov2017_ICLR,
  title={Decoupled Weight Decay Regularization},
  author={Loshchilov, Ilya and Hutter, Frank},
  booktitle={International Conference on Learning Representations}
}

@inproceedings{Wofk2023_ICRA,
  title={Monocular visual-inertial depth estimation},
  author={Wofk, Diana and Ranftl, Ren{\'e} and M{\"u}ller, Matthias and Koltun, Vladlen},
  booktitle={2023 IEEE International Conference on Robotics and Automation (ICRA)},
  pages={6095--6101},
  year={2023},
  organization={IEEE}
}

@article{Wong2020_RAL,
  title={Unsupervised depth completion from visual inertial odometry},
  author={Wong, Alex and Fei, Xiaohan and Tsuei, Stephanie and Soatto, Stefano},
  journal={IEEE Robotics and Automation Letters},
  volume={5},
  number={2},
  pages={1899--1906},
  year={2020},
  publisher={IEEE}
}

@inproceedings{Teed2020_ECCV,
  title={Raft: Recurrent all-pairs field transforms for optical flow},
  author={Teed, Zachary and Deng, Jia},
  booktitle={European conference on computer vision},
  pages={402--419},
  year={2020},
  organization={Springer}
}

@article{Gu2023_RAL,
  title={DRO: Deep recurrent optimizer for video to depth},
  author={Gu, Xiaodong and Yuan, Weihao and Dai, Zuozhuo and Zhu, Siyu and Tang, Chengzhou and Dong, Zilong and Tan, Ping},
  journal={IEEE Robotics and Automation Letters},
  volume={8},
  number={5},
  pages={2844--2851},
  year={2023},
  publisher={IEEE}
}

@InProceedings{Rosinol2020_Kimera,
  title = {Kimera: an Open-Source Library for Real-Time Metric-Semantic Localization and Mapping},
  author = {Rosinol, Antoni and Abate, Marcus and Chang, Yun and Carlone, Luca},
  year = {2020},
  booktitle = {IEEE Intl. Conf. on Robotics and Automation (ICRA)},
  url = {https://github.com/MIT-SPARK/Kimera}
}

@inproceedings{Wong2021_ICCV,
  title={Unsupervised depth completion with calibrated backprojection layers},
  author={Wong, Alex and Soatto, Stefano},
  booktitle={Proceedings of the IEEE/CVF International Conference on Computer Vision},
  pages={12747--12756},
  year={2021}
}

@article{Zhao2022_MDPI,
  title={Dit-slam: real-time dense visual-inertial slam with implicit depth representation and tightly-coupled graph optimization},
  author={Zhao, Mingle and Zhou, Dingfu and Song, Xibin and Chen, Xiuwan and Zhang, Liangjun},
  journal={Sensors},
  volume={22},
  number={9},
  pages={3389},
  year={2022},
  publisher={MDPI}
}

@inproceedings{Zuo2021_ICRA,
  title={CodeVIO: Visual-inertial odometry with learned optimizable dense depth},
  author={Zuo, Xingxing and Merrill, Nathaniel and Li, Wei and Liu, Yong and Pollefeys, Marc and Huang, Guoquan},
  booktitle={2021 ieee international conference on robotics and automation (icra)},
  pages={14382--14388},
  year={2021},
  organization={IEEE}
}

@article{Merrill2024_arxiv,
  title={Visual-inertial SLAM as simple as A, B, VINS},
  author={Merrill, Nathaniel and Huang, Guoquan},
  journal={arXiv preprint arXiv:2406.05969},
  year={2024}
}

@article{Ranftl2021_ICCV,
	author    = {Ren\'{e} Ranftl and Alexey Bochkovskiy and Vladlen Koltun},
	title     = {Vision Transformers for Dense Prediction},
	journal   = {ICCV},
	year      = {2021},
}

@article{Birkl2023_arxiv,
      title={MiDaS v3.1 -- A Model Zoo for Robust Monocular Relative Depth Estimation},
      author={Reiner Birkl and Diana Wofk and Matthias M{\"u}ller},
      journal={arXiv preprint arXiv:2307.14460},
      year={2023}
}

@article{Bhat2023_arxiv,
  title={Zoedepth: Zero-shot transfer by combining relative and metric depth},
  author={Bhat, Shariq Farooq and Birkl, Reiner and Wofk, Diana and Wonka, Peter and M{\"u}ller, Matthias},
  journal={arXiv preprint arXiv:2302.12288},
  year={2023}
}

@article{Fei2019_RAL,
  title={Geo-supervised visual depth prediction},
  author={Fei, Xiaohan and Wong, Alex and Soatto, Stefano},
  journal={IEEE Robotics and Automation Letters},
  volume={4},
  number={2},
  pages={1661--1668},
  year={2019},
  publisher={IEEE}
}

@inproceedings{Yang2017_ICRA,
  title     = {Real-time Monocular Dense Mapping on Aerial Robots using Visual-Inertial Fusion},
  author    = {Yang, Zhenfei and Gao, Fei and Shen, Shaojie},
  booktitle = {2017 IEEE International Conference on Robotics and Automation (ICRA)},
  pages     = {4552--4559},
  year      = {2017},
  organization = {IEEE},
  doi       = {10.1109/ICRA.2017.7989529}
}

@article{Merrill2024_IJRR,
  title={Fast and robust learned single-view depth-aided monocular visual-inertial initialization},
  author={Merrill, Nathaniel and Geneva, Patrick and Katragadda, Saimouli and Chen, Chuchu and Huang, Guoquan},
  journal={The International Journal of Robotics Research},
  pages={02783649241262452},
  publisher={SAGE Publications Sage UK: London, England}
}

@article{Merrill2023_RSS,
  title={Fast monocular visual-inertial initialization leveraging learned single-view depth},
  journal={Robotics: Science and Systems (RSS) 2023},
  year={2023},
  publisher={IEEE}
}

@article{Godard2019_ICCV_monodepth2,
  title     = {Digging into Self-Supervised Monocular Depth Prediction},
  author    = {Cl{\'{e}}ment Godard and
               Oisin {Mac Aodha} and
               Michael Firman and
               Gabriel J. Brostow},
  booktitle = {The International Conference on Computer Vision (ICCV)},
  month = {October},
year = {2019}
}

@inproceedings{Mourikis2007_ICRA,
  title={A multi-state constraint Kalman filter for vision-aided inertial navigation},
  author={Mourikis, Anastasios I and Roumeliotis, Stergios I},
  booktitle={Proceedings 2007 IEEE international conference on robotics and automation},
  pages={3565--3572},
  year={2007},
  organization={IEEE}
}

@inproceedings{Geneva2020ICRA,
	title        = {{OpenVINS:} A Research Platform for Visual-Inertial Estimation},
	author       = {Patrick Geneva and Kevin Eckenhoff and Woosik Lee and Yulin Yang and Guoquan Huang},
	year         = 2020,
	booktitle    = {Proc. of the IEEE International Conference on Robotics and Automation},
	address      = {Paris, France},
	url          = {https://github.com/rpng/open_vins},
}

@inproceedings{Huang2019ICRA,
	title        = {Visual-Inertial Navigation: A Concise Review},
	author       = {Guoquan Huang},
	year         = 2019,
	month        = may,
	booktitle    = {Proc. International Conference on Robotics and Automation},
	address      = {Montreal, Canada},
}

@inproceedings{Zuo2021ICRA,
	title        = {CodeVIO: Visual-Inertial Odometry with Learned Optimizable Dense Depth},
	author       = {Xingxing Zuo and Nate Merrill and Wei Li and Yong Liu and Marc Pollefeys and Guoquan Huang},
	year         = 2021,
	booktitle    = {Proc. of the IEEE International Conference on Robotics and Automation},
	address      = {Xi'an, China},
}

@InProceedings{Merrill2023RSS,
  author    = {Nate Merrill and Patrick Geneva and Saimouli Katragadda and Chuchu Chen and Guoquan Huang},
  booktitle = {Proc. Robotics: Science and Systems (RSS)},
  title     = {Fast Monocular Visual-Inertial Initialization Leveraging Learned Single-View Depth},
  year      = {2023},
  address   = {Daegu, Republic of Korea},
  month     = jul,
}

@InProceedings{Chen2023ICRA,
  author    = {Chuchu Chen and Patrick Geneva and Yuxiang Peng and Woosik Lee and Guoquan Huang},
  booktitle = {Proc. of the IEEE International Conference on Robotics and Automation},
  title     = {Monocular Visual-Inertial Odometry with Planar Regularitie},
  year      = {2023},
  address   = {London, UK.},
}

@Article{Yang2023TRO,
  author    = {Yulin Yang and Patrick Geneva and Xingxing Zuo and Guoquan Huang},
  journal   = {IEEE Transactions on Robotics},
  title     = {Online Self-Calibration for Visual-Inertial Navigation Systems: Models, Analysis and Degeneracy},
  year      = {2023},
  month     = may,
}

@Conference{Katragadda2025ICCV,
  author           = {Saimouli Katragadda and Cho-Ying Wu and Yuliang Guo and Xinyu Huang and Guoquan Huang and Liu Ren},
  booktitle        = {International Conference on Computer Vision (ICCV)},
  title            = {Online Language Splatting},
  year             = {2025},
  archiveprefix    = {arXiv},
  creationdate     = {2025-08-12T11:09:46},
  eprint           = {2503.09447},
  modificationdate = {2025-08-12T11:11:15},
  primaryclass     = {cs.AI},
  url              = {https://arxiv.org/abs/2503.09447},
}
}

\end{document}